\pdfoutput=1

\documentclass[11pt]{article}

\usepackage{authblk}

\usepackage{acl}

\usepackage{times}
\usepackage{latexsym}

\usepackage[T1]{fontenc}

\usepackage[utf8]{inputenc}

\usepackage{microtype}

\usepackage{graphicx}
\usepackage{amsmath}
\usepackage{bm}
\usepackage{multirow}
\usepackage{booktabs}
\usepackage{subcaption}
\usepackage{CJKutf8}
\usepackage{caption}
\usepackage{amssymb}
\usepackage{stfloats}
\usepackage{ulem}
\title{TranS: Transition-based Knowledge Graph Embedding \\ with Synthetic Relation Representation}
\author{{\bf Xuanyu Zhang}, {\bf Qing Yang} \and {\bf Dongliang Xu} \\
        DXM-DI-AI
        }

\begin{document}
\maketitle
\begin{abstract}
Knowledge graph embedding (KGE) aims to learn continuous vectors of relations and entities in knowledge graph. Recently, transition-based KGE methods have achieved promising performance, where the single relation vector learns to translate head entity to tail entity. However, this scoring pattern is not suitable for complex scenarios where the same entity pair has different relations. Previous models usually focus on the improvement of entity representation for 1-to-N, N-to-1 and N-to-N relations, but ignore the single relation vector. In this paper, we propose a novel transition-based method, TranS, for knowledge graph embedding. The single relation vector in traditional scoring patterns is replaced with synthetic relation representation, which can solve these issues effectively and efficiently. Experiments on a large knowledge graph dataset, ogbl-wikikg2, show that our model achieves state-of-the-art results.
\end{abstract}

\section{Introduction}
Knowledge graphs (KGs), such as Freebase \cite{bollacker2008freebase}, Yago \cite{rebele2016yago} and Wikidata \cite{vrandevcic2014wikidata}, play an important role in many applications, including question answering, information retrieval and so on. KG is usually represented as the form of triplets (\textit{head entity}, \textit{relation}, \textit{tail entity})(denoted as (\textit{h}, \textit{r}, \textit{t})), where \textit{relation} 
indicates the relationship between the two entities. With the rapid development of deep learning,
many representation learning methods \cite{bordes2013translating,wang2014knowledge,fan2014transition,lin2015learning,ji2015knowledge,ji2016knowledge,xie2017interpretable,qian2018translating,chao2020pairre,long2021triplere,chen2021relation,wang2022interht} are proposed to obtain  low-dimensional embedding vectors of entities and relations in KG.

\begin{figure}[t]
\begin{center}
\includegraphics[width=0.5\textwidth]{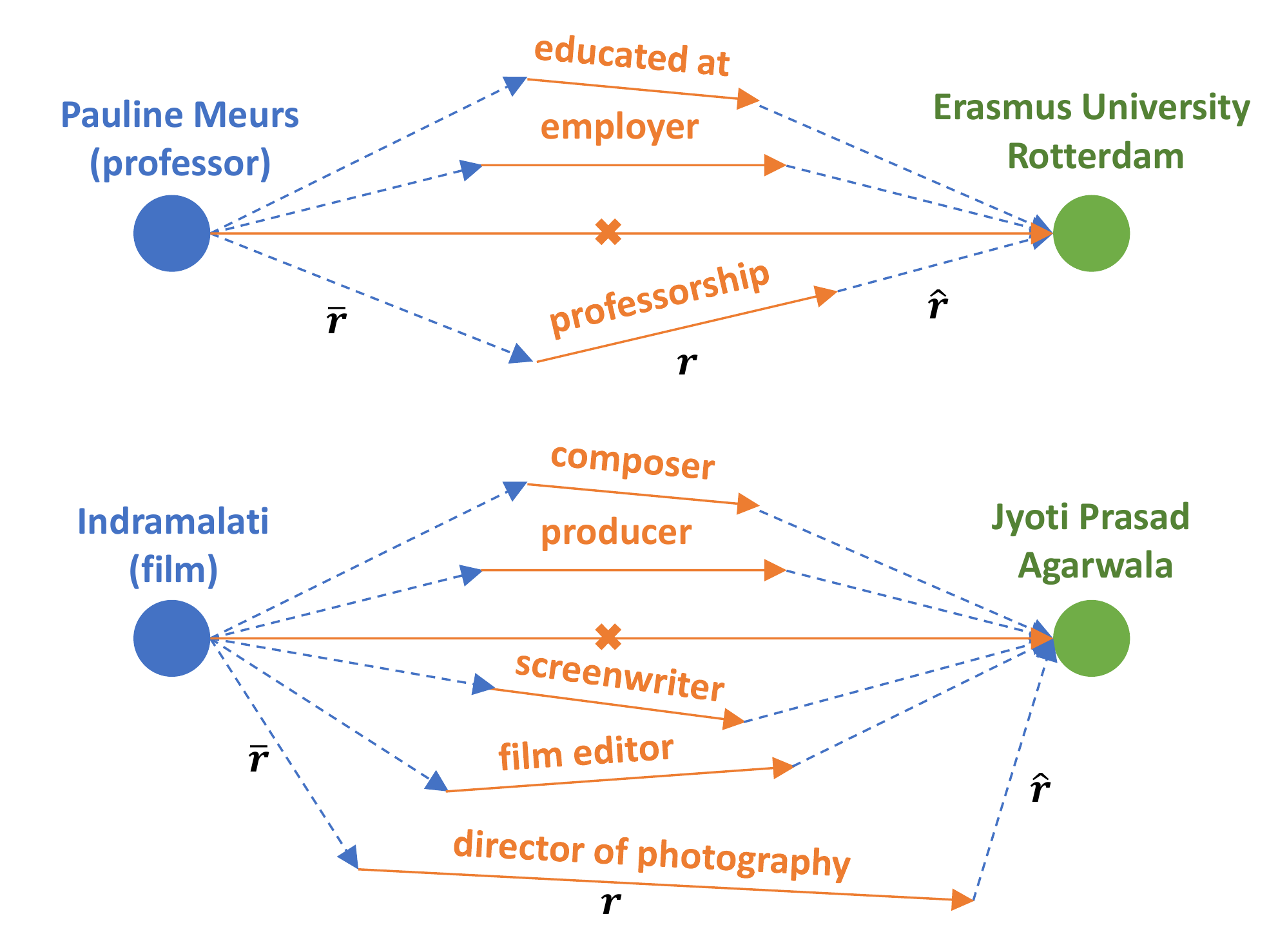}
\caption{Examples from ogbl-wikikg2. It is difficult for single relation vector to represent different relations between the same entity pairs.
}\label{intro}
\end{center}
\end{figure}

Generally speaking, knowledge graph embedding (KGE) methods can be roughly divided into the following directions: translational distance \cite{bordes2013translating,wang2014knowledge,fan2014transition,lin2015learning,ji2015knowledge}, semantic matching \cite{10.5555/3104482.3104584,10.5555/3120260.3120289,yang2014embedding,10.5555/3016100.3016172,trouillon2016complex} and neural networks \cite{10.1007/s10994-013-5363-6,DBLP:journals/corr/Liu0LWH16,li2016gated, 10.5555/3305381.3305512, xu2018how}.
Because transition-based KGE method like TransE \cite{bordes2013translating} is simple but effective, this series of models are becoming more and more popular in both academia and industry. Specifically, TransE makes the difference between two entity vectors ($\bf{h}$ and $\bf{t}$) approximate to the relation vector ($\bf{r}$), i.e., $\bf{t}-h \approx r$. That is to say, the relation \textit{r} is characterized by the translating vector $\bf{r}$.
However, TransE is not suitable to deal with complex relations like one-to-many/many-to-one/many-to-many. 
Especially, the same entity pair usually has many different relations. For example, in Figure \ref{intro}, after graduating from Erasmus University Rotterdam, the professor became a professor of the same university.
The composer, producer, screenwriter, editor and director of the film, Indramalati, can be the same person, Jyoti Prasad Agarwala. As shown in Table \ref{tab:related}, although previous models like TransH/R/D \cite{wang2014knowledge,lin2015learning,ji2015knowledge} solved the problem of 1-to-N, N-to-1 and N-to-N, they still continued the TransE pattern, $ \bf{R_t}  - \bf{R_h}  \approx \bf{r}$, where $ \bf{R_t}$ and $ \bf{R_h}$ are the deformation of  $ \bf{t}$ and $ \bf{h}$ respectively. Even if the entity vector is represented by hyperplane or multiple embedding spaces, single relation vector $\bf{r}$ still cannot represent different relationships when facing the same entity pair.

To solve these issues, we propose a novel transition-based knowledge graph embedding model, TranS, which replaces tradition scoring pattern with synthetic relation pattern, i.e., $\bf{R_t}  - \bf{R_h} \approx \bar{r}+r+\hat{r}$. The final relation representation is the sum of multiple relation vectors. It can handle not only complex relations in knowledge graph, but also the situation in Figure \ref{intro}, where orange solid lines denote $\bf{r}$, and  blue dotted lines denote $\bf{\bar{r}},\bf{\hat{r}}$. Experiments on a large knowledge graph dataset, ogbl-wikikg2, show that our proposed model achieves  state-of-the-art results.

\begin{table*}[!ht]
\small
\begin{center}
\begin{tabular}{l l l l }
\toprule
Model & Embedding & Scoring Function & Scoring Pattern \\
\midrule
TransE&	$\mathbf{h}, \mathbf{t} \in \mathbb{R}^{d}$, 	$\mathbf{r} \in \mathbb{R}^{d}$	&	$-\|\mathbf{h}+\mathbf{r}-\mathbf{t}\|_{1 / 2}$ & $|| \bf{R_h}  - \bf{R_t}  + \bf{r}||$\\
TransR	&	$\mathbf{h}, \mathbf{t} \in \mathbb{R}^{d}$	, 	$\mathbf{r} \in \mathbb{R}^{k}, \mathbf{M}_{r} \in \mathbb{R}^{k \times d}$	&	$-\left\|\mathbf{M}_{r} \mathbf{h}+\mathbf{r}-\mathbf{M}_{r} \mathbf{t}\right\|_{2}^{2}$ & $|| \bf{R_h}  - \bf{R_t}  + \bf{r}||$\\
TransH	&	$\mathbf{h}, \mathbf{t} \in \mathbb{R}^{d}$, 	$\mathbf{r}, \mathbf{w}_{r} \in \mathbb{R}^{d}$	&	$-\left\|\left(\mathbf{h}-\mathbf{w}_{r}^{\top} \mathbf{h} \mathbf{w}_{r}\right)+\mathbf{r}-\left(\mathbf{t}-\mathbf{w}_{r}^{\top} \mathbf{t} \mathbf{w}_{r}\right)\right\|_{2}^{2}$ & $|| \bf{R_h}  - \bf{R_t}  + \bf{r}||$\\
ITransF	&	 $\mathbf{h}, \mathbf{t} \in \mathbb{R}^d$,	$\mathbf{r}\in\mathbb{R}^d$	&	$\left\|\boldsymbol{\alpha}_{r}^{H} \cdot \mathbf{D} \cdot \mathbf{h}+\mathbf{r}-\boldsymbol{\alpha}_{r}^{T} \cdot \mathbf{D} \cdot \mathbf{t}\right\|_{\ell}$ 	 & $|| \bf{R_h}  - \bf{R_t}  + \bf{r}||$\\
TransAt	&	$\mathbf{h}, \mathbf{t} \in \mathbb{R}^{d}$,	$\mathbf{r} \in \mathbb{R}^{d}$	&	$P_{r}\left(\sigma\left(\mathbf{r}_{h}\right) \mathbf{h}\right)+\mathbf{r}-P_{r}\left(\sigma\left(\mathbf{r}_{t}\right) \mathbf{t}\right)$	 & $|| \bf{R_h}  - \bf{R_t}  + \bf{r}||$\\
TransD	&	$\mathbf{h}, \mathbf{t}, \mathbf{w}_{h} \mathbf{w}_{t} \in \mathbb{R}^{d}$,	$\mathbf{r}, \mathbf{w}_{r} \in \mathbb{R}^{k}$	&	$-\left\|\left(\mathbf{w}_{r} \mathbf{w}_{h}^{\top}+\mathbf{I}\right) \mathbf{h}+\mathbf{r}-\left(\mathbf{w}_{r} \mathbf{w}_{t}^{\top}+\mathbf{I}\right) \mathbf{t}\right\|_{2}^{2}$	 & $|| \bf{R_h}  - \bf{R_t}  + \bf{r}||$\\
TransM	&	$\mathbf{h}, \mathbf{t} \in \mathbb{R}^{d}$,	$\mathbf{r} \in \mathbb{R}^{d}$	&	$-\theta_{r}\|\mathbf{h}+\mathbf{r}-\mathbf{t}\|_{1 / 2}$	 & $|| \bf{R_h}  - \bf{R_t}  + \bf{r}||$\\
\multirow{2}{30pt}{TranSparse}	&$\mathbf{h}, \mathbf{t} \in \mathbb{R}^{d}$,	$\mathbf{r} \in \mathbb{R}^{k}, \mathbf{M}_{r}\left(\theta_{r}\right) \in \mathbb{R}^{k \times d}$	&	$-\left\|\mathbf{M}_{r}\left(\theta_{r}\right) \mathbf{h}+\mathbf{r}-\mathbf{M}_{r}\left(\theta_{r}\right) \mathbf{t}\right\|_{1 / 2}^{2}$	 & $|| \bf{R_h}  - \bf{R_t}  + \bf{r}||$\\
&	$\mathbf{h}, \mathbf{t} \in \mathbb{R}^{d}$,$\mathbf{M}_{r}^{1}\left(\theta_{r}^{1}\right), \mathbf{M}_{r}^{2}\left(\theta_{r}^{2}\right) \in \mathbb{R}^{k \times d}$	&	$-\left\|\mathbf{M}_{r}^{1}\left(\theta_{r}^{1}\right) \mathbf{h}+\mathbf{r}-\mathbf{M}_{r}^{2}\left(\theta_{r}^{2}\right) \mathbf{t}\right\|_{1 / 2}^{2}$ & $|| \bf{R_h}  - \bf{R_t}  + \bf{r}||$\\
\midrule
\multirow{2}{30pt}{TranS} 	&\multirow{2}{150pt}{$\mathbf{h}, \mathbf{t}, \mathbf{\tilde{h}}, \mathbf{\tilde{t}} \in \mathbb{R}^{d}$, 	$\mathbf{r}, \bf{\bar{r}}, \bf{\hat{r}} \in \mathbb{R}^{d}$} 		&	\multirow{2}{100pt}{$ -||\bf{h} \circ \tilde{t} - \bf{t} \circ \tilde{h} + \uwave{\bf{\bar{r}} \circ \bf{h} + \bf{r} +  \bf{\hat{r}} \circ  \bf{t} } ||$} & $|| \bf{R_h}  - \bf{R_t} + $\\
&&&$ \bf{\bar{r}}+r+\hat{r}||$\\
\bottomrule 
\end{tabular}
\caption{Transition-based knowledge graph embedding models.}\label{tab:related}
\end{center}
\end{table*}

\section{Methodology}
In this section, we will first introduce  our proposed TranS model. 
And we then introduce the loss function during training and differences among different KGE methods.

\subsection{TranS}
\begin{figure*}[t]
\begin{center}
\includegraphics[width=\textwidth]{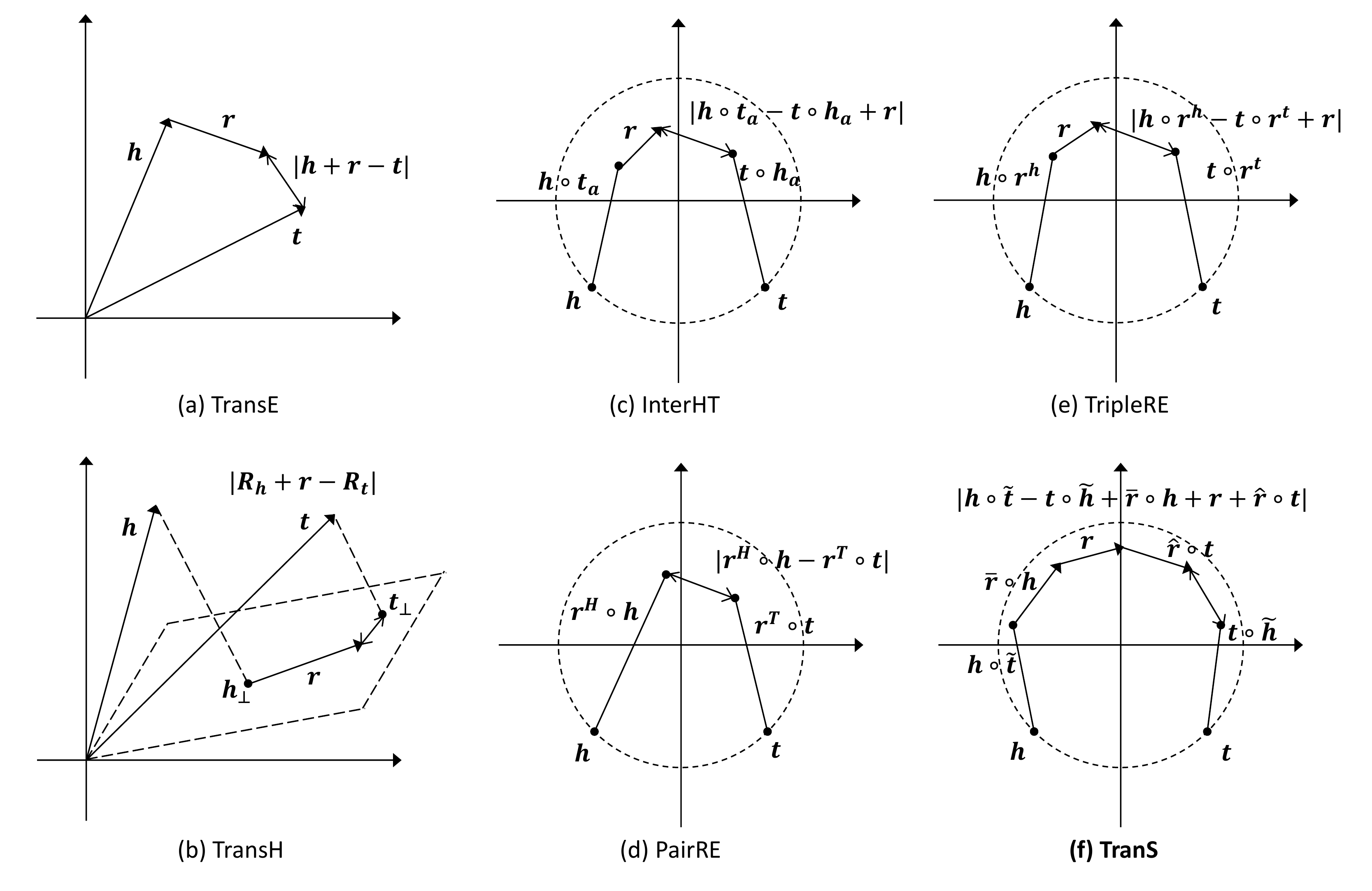}
\caption{Comparison of different  transition-based KGE models.
}\label{transs}
\end{center}
\end{figure*}

Our proposed TranS model first breaks the traditional scoring patterns  $\bf{R_t}  - \bf{R_h} \approx r$  in previous models \cite{bordes2013translating,wang2014knowledge,fan2014transition,lin2015learning,chao2020pairre,long2021triplere,wang2022interht}. It replaces single relation vector $\bf{r}$ with synthetic relation vectors $\bf{\bar{r}}+r+\bf{\hat{r}}$, where $\bf{\bar{r}},\bf{\hat{r}}$ are related to head and tail entities.
That is to say, we want  $\bf{R_t}  - \bf{R_h} \approx \bar{r}+r+\hat{r}$.
The illustration of TranS is shown in Figure \ref{transs} (f). 
Two entity and three relation representations together make up our proposed scoring function $f_r(h,t)$.
Especially,  the synthetic relation representation consists of the sum of three different relation vectors.
To make full use of context information, we use combined vectors with Hadamard product to represent $\bf{h}$, $\bf{t}$, $\bf{\bar{r}}$ and $\bf{\hat{r}}$ separately:
\begin{equation}
\begin{aligned} 
f_r(h,t) &= -|| \bf{R_h}  - \bf{R_t}  + \bf{R_r}||\\
\bf{R_h} &= \bf{h} \circ \tilde{t} \\
\bf{R_t} &= \bf{t} \circ \tilde{h} \\
\bf{R_r} &= \bf{\bar{r}} \circ \bf{h} + \bf{r} +  \bf{\hat{r}} \circ  \bf{t} 
\end{aligned}
\end{equation}
where $\mathbf{h}$, $\mathbf{t}$ and $\mathbf{r}$ denote main vectors similar to traditional scoring patterns.
$\mathbf{\tilde{h}}$ is auxiliary head entity vector and $\mathbf{\tilde{t}}$ is auxiliary tail entity vector.
And $\mathbf{\bar{r}}$ is auxiliary relation vector related to head entity and $\mathbf{\hat{r}}$ is auxiliary relation vector related to tail entity.
$\bf{R_h}$ is the representation of the head that combines information of the tail, and $\bf{R_t}$ is the representation of the tail integrating information of the head. 
$\bf{\bar{r}} \circ \bf{h}$  is the representation of the relation that combines information of the head, and $\bf{\hat{r}} \circ  \bf{t}$  is the representation of another relation integrating information of the tail. 
Thus the final equations can be represented as follows:
\begin{equation}
\begin{aligned} 
-|| \bf{h} \circ \tilde{t} - \bf{t} \circ \tilde{h} + \uwave{\bf{\bar{r}} \circ \bf{h} + \bf{r} +  \bf{\hat{r}} \circ  \bf{t} }||
\end{aligned} \label{eq:finaleq}
\end{equation}

Following previous works \cite{long2021triplere,wang2022interht}, we add an unit vector $  \bf{e}$ to $\bf{R_h}$ and $\bf{R_t}$, i.e., $ \bf{h} \circ \tilde{t} \rightarrow \bf{h} \circ (\tilde{t} + e)$, $ \bf{t} \circ \tilde{h}  \rightarrow \bf{t} \circ (\tilde{h}+ e)$. And considering the OOV problem, we also use the NodePiece \cite{galkin2021nodepiece} to learn a fixed-size entity vocabulary.

\subsection{Training}
In the training process, we use the self-adversarial negative sampling loss \cite{sun2019rotate} as the loss function. The loss function is as follows:
\begin{equation}
\begin{aligned}
 \mathcal{L} = & -log\ \sigma(\gamma-d_r(\mathbf{h},\mathbf{t}))\\& -\sum_{i=1}^{n} \frac{1}{k} log\ \sigma (d_r (\mathbf{h_i '}, \mathbf{t_i '})-\gamma )
\end{aligned}
\end{equation}
where $\gamma$ is a fixed margin, $\sigma$ is the sigmoid function, and $(\mathbf{h_i '}, \mathbf{r}, \mathbf{t_i '})$ is the $i$-th negative triplet.

\subsection{Comparison}
As shown in Figure \ref{transs}, the main difference between our model (f) and previous transition-based KGE methods (a,b,c,d,e) is the synthetic relation representation.
That is to say, it changes single relation representation $\bf{r}$ in tradition scoring pattern $\bf{R_t}  - \bf{R_h} \approx r$ to synthetic relation representation $\bf{\bar{r}}+r+\hat{r}$ in our proposed new pattern  $\bf{R_t}  - \bf{R_h} \approx \bar{r}+r+\hat{r}$.
Specifically,
different from InterHT \cite{wang2022interht}, the relation part of our scoring function is the sum of multiple relation vectors $\bf{R_r} = \bf{\bar{r}} \circ \bf{h} + \bf{r} +  \bf{\hat{r}} \circ  \bf{t} $ rather than single vector $ \bf{r} $.
Comparing with TripleRE \cite{long2021triplere}, where three relations are applied into three parts  ($\bf{R_h} = \bf{h} \circ r^h$, $\bf{R_t} = \bf{t} \circ r^t$ , $\bf{R_r} =  \bf{r^m}$) of traditional scoring patterns with addition and subtraction operations, our proposed TranS only applies synthetic relation vectors into the relation part $\bf{R_r} = \bf{\bar{r}} \circ \bf{h} + \bf{r} +  \bf{\hat{r}} \circ  \bf{t} $ of scoring functions with vector addition operation.

\section{Experiments}
In this section, we will first introduce the dataset and metric.
Then we will  introduce  implementation details and experimental results further.

\begin{table*}[!tp]
\begin{center}
\scalebox{1.}{
\begin{tabular}{lcccc}
\toprule
Model & \#Params & \#Dims & Test MRR & Valid MRR \\
\midrule
TransE & 1251M & 500 & 0.4256 $\pm$ 0.0030 & 0.4272 $\pm$ 0.0030 \\
RotatE & 1250M & 250 & 0.4332 $\pm$ 0.0025 & 0.4353 $\pm$ 0.0028 \\
PairRE & 500M & 200 & 0.5208 $\pm$ 0.0027 & 0.5423 $\pm$ 0.0020 \\
AutoSF & 500M & - & 0.5458 $\pm$ 0.0052 & 0.5510 $\pm$ 0.0063 \\
ComplEx & 1251M & 250 & 0.5027 $\pm$ 0.0027 & 0.3759 $\pm$ 0.0016 \\
TripleRE & 501M & 200 & 0.5794 $\pm$ 0.0020 & 0.6045 $\pm$ 0.0024 \\
\midrule
ComplEx-RP & 250M & 50 & 0.6392 $\pm$ 0.0045 & 0.6561 $\pm$ 0.0070 \\
AutoSF + NodePiece & 6.9M & - & 0.5703 $\pm$ 0.0035 & 0.5806 $\pm$ 0.0047 \\
TripleREv2 + NodePiece & 7.3M & 200 & 0.6582 $\pm$ 0.0020 & 0.6616 $\pm$ 0.0018 \\
InterHT + NodePiece & 19.2M & 200 & 0.6779 $\pm$ 0.0018 & 0.6893 $\pm$ 0.0015  \\
TripleREv3 + NodePiece & 36.4M & 200 & 0.6866 $\pm$ 0.0014 & 0.6955 $\pm$ 0.0008 \\
\textbf{TranS + NodePiece} & \textbf{19.2M} & 200 & \textbf{0.6882 $\pm$ 0.0019} & \textbf{0.6988 $\pm$ 0.0006} \\
\bottomrule
\end{tabular}}
\end{center}
\caption{Results on the ogbl-wikikg2 dataset.}\label{experiment_results}
\end{table*}

\subsection{Dataset and Metric}

The ogbl-wikikg2 \cite{hu2020ogb} is a large KG dataset extracted from Wikidata \cite{vrandevcic2014wikidata}.
It contains a set of triplet edges, capturing the different types of relations between entities in the world. 
The statistics of the dataset are show in Table \ref{dataset}. It contains 2,500,604 entities, 535 relation types and 17,137,181 edges. Following official guidelines, we evaluate the KGE performance by predicting new triplet edges according to the training edges. The evaluation metric follows the standard filtered metric widely used in KG. 
Specifically, each test triplet edges are corrupted by replacing its head or tail with randomly-sampled 1,000 negative entities, while ensuring the resulting triplets do not appear in KG. The goal is to rank the true head (or tail) entities higher than the negative entities, which is measured by Mean Reciprocal Rank (MRR).

We use original dataset splitting configuration. It splits the triplets according to time, simulating a realistic KG completion scenario that aims to fill in missing triplets that are not present at a certain timestamp. Specifically, the training set contains 16,109,182 triplets, the validation set contains 429,456 triplets, and the test set contains 598,543 triplets.

\begin{table}[!tp]
\begin{center}
\begin{tabular}{ll}
\toprule
Data & \#Number \\
\midrule
Nodes & 2,500,604 \\
Relations & 535 \\
Edges & 17,137,181 \\
\midrule
Train & 16,109,182 \\
Validation & 429,456 \\
Test & 598,543 \\
\bottomrule
\end{tabular}
\end{center}
\caption{Statistics of the ogbl-wikikg2 dataset.}\label{dataset}
\end{table}

\subsection{Implementation Details}
In our experiments, the Adam \cite{kingma2014adam} is used as our optimizer with 0.0005 learning rate. The batch size of the model is set to 512. To prevent overfitting, we use dropout technique, and set it to 0.1. The negative sampling size  is set to 128. And the dimension of each embedding vector in Eq. \ref{eq:finaleq} is set to 200. The maximum number of training steps is 800 thousand. We validate the model every 20 thousand steps. The number of anchors for NodePiece are 20 thousand. And $\gamma$ in the loss function is set to 6. These hyper-parameters are selected according to the performance on the validation set.

\subsection{Results}
The results are shown in Table \ref{experiment_results}. Our model achieves 0.6988 (validation set) and 0.6882 (test set) on MRR, which outperforms previous best model, TripleREv3, on the ogbl-wikikg2 dataset. Especially, the parameters of our model (19.2M) are about half of TripleREv3 (36.4M). The experimental results show that our proposed method can improve the model performance effectively with fewer parameters. Besides, we also construct the 38.4M TranS (large) model, the best score of which can reach 0.7101 (validation set) and 0.6992 (test set) on MRR.

\section{Conclusion}
In this paper, we propose a novel transition-based knowledge graph embedding model, TranS, to solve the issues of complex relations, especially the situation of same entity pair with different relations. TranS replaces tradition scoring pattern with synthetic relation pattern. And the final relation representation is the sum of different relation vectors. Experiments on the ogbl-wikikg2 dataset show that our model achieves  state-of-the-art results.

\bibliography{anthology}
\bibliographystyle{acl_natbib}

\end{document}